\documentclass[letterpaper, 10 pt, conference]{Format/ieeeconf}  

\IEEEoverridecommandlockouts                              

\usepackage{microtype}

\usepackage{graphicx}
\usepackage{amsmath}
\usepackage{amssymb}
\usepackage{algorithm}
\usepackage{algpseudocode}
\usepackage{cite}
\usepackage{booktabs}
\usepackage{caption}
\usepackage{multirow}
\usepackage{xcolor}
\algrenewcommand\algorithmicrequire{\textbf{Input:}}
\algrenewcommand\algorithmicensure{\textbf{Output:}}
\usepackage{array}

\title{\LARGE \bf
DynaFlow: Dynamics-embedded Flow Matching for Physically Consistent Motion Generation from State-only Demonstrations
}

\author{
Sowoo Lee$^{*1}$, 
Dongyun Kang$^{*1}$,
Jaehyun Park$^{*1}$
and Hae-Won Park$^{1}$%
\thanks{This work was supported by the Robotics and AI Institute.}
\thanks{$^{*}$Equal contribution.}
\thanks{
$^{1}$Korea Advanced Institute of Science and Technology, Yuseong-gu, Daejeon 34141, Republic of Korea. {\tt\small haewonpark@kaist.ac.kr}}
\thanks{Paper website: \textcolor{blue}{https://sowoolee.github.io/dynaflow}}
}

\begin{document}

\maketitle
\thispagestyle{empty}
\pagestyle{empty}

\begin{abstract}

This paper introduces DynaFlow, a novel framework that embeds a differentiable simulator directly into a flow matching model. By generating trajectories in the action space and mapping them to dynamically feasible state trajectories via the simulator, DynaFlow ensures all outputs are physically consistent by construction. This end-to-end differentiable architecture enables training on state-only demonstrations, allowing the model to simultaneously generate physically consistent state trajectories while inferring the underlying action sequences required to produce them.
We demonstrate the effectiveness of our approach through quantitative evaluations and showcase its real-world applicability by deploying the generated actions onto a physical Go1 quadruped robot. 
The robot successfully reproduces diverse gaits present in the dataset, executes long-horizon motions in open-loop control and translates infeasible kinematic demonstrations into dynamically executable stylistic behaviors. These hardware experiments validate that DynaFlow produces deployable, highly effective motions on real-world hardware from state-only demonstrations, effectively bridging the gap between kinematic data and real-world execution.

\end{abstract}


\section{Introduction}

Generative models, such as Diffusion Models and Flow Matching, have recently achieved unprecedented success across various domains, including image\cite{ddpm_2020, song2020score, ldm_2022}, audio\cite{diffwave_2020}, and text generation\cite{diffusionLM_2022}. They have demonstrated a remarkable ability to learn intricate data distributions from large-scale datasets, producing highly natural and diverse outputs. Inspired by this success, these models are increasingly being recognized as powerful tools for generating complex motion trajectories in fields like robotics and computer graphics\cite{mdm_2022, diffuser_2022, diffusionpolicy_2023}. 
Indeed, their application to kinematic motion generation has seen significant progress, largely driven by the increasing availability of state demonstration data from sources such as motion capture and raw video.

However, directly applying these generative models to character animation and robot control presents significant challenges. 
A primary limitation is the lack of physical consistency.
Most generative models learn a statistical approximation of the data distribution from a finite set of examples, rather than the underlying physical principles governing the data. Consequently, there is no guarantee that the generated outputs will adhere to intrinsic physical principles or dynamic constraints. 
This means that when generating novel behaviors, the resulting motions can be either physically implausible, with artifacts like ground penetration, character floating, and foot sliding, or dynamically inconsistent, making them unsuitable to execute in the physical world.
The issue becomes particularly pronounced when models are trained on datasets with inherent physical inconsistencies, such as those sourced from motion capture or generated via kinematic retargeting.

\textcolor{black}{
Another major hurdle is the scarcity of action data. While kinematic state trajectories are abundant, the corresponding hardware-specific action sequences (e.g., joint torques), necessary for deployment, are costly to collect. To bridge this gap, prior diffusion-based control approaches commonly employ a hierarchical framework, where the generative model produces desired state trajectories that are subsequently executed by a low-level tracking controller.
Since dynamic feasibility is not enforced at the generative stage, stable execution hinges on additional adaptation between the planner and controller.
Despite various efforts, mitigating this mismatch remains a significant challenge~\cite{diffusecloc_2025}.
An alternative strategy follows a multi-expert distillation paradigm, where skill-specific expert controllers are first trained and their rollouts are distilled into a unified generative model. However, acquiring high-quality action experts is itself cumbersome, particularly when scaling to a diverse set of skills. Furthermore, the distillation stage introduces additional data engineering overhead to fully capture the robustness of these experts.
}

To overcome these limitations, we propose DynaFlow, a novel dynamics-embedded flow matching framework that guarantees physically consistent motion generation while 
inferring deployable actions directly from state-only demonstrations.
The core idea of DynaFlow is to integrate a differentiable simulator at the output of the flow matching prediction module. This simulator layer acts as a mapping from the space of action trajectories to the space of dynamically feasible state trajectories, ensuring that the model's output strictly adheres to the laws of physics by construction. Furthermore, its differentiable nature enables end-to-end training via analytical gradients. During optimization, the model naturally discovers the action trajectory required to reconstruct a given state trajectory in a dynamically consistent manner.
\textcolor{black}{Consequently, DynaFlow 
alleviates the need for auxiliary tracking controllers or complex multi-stage distillation pipelines.}

We conduct
experiments to validate the effectiveness of DynaFlow. Our quantitative analysis compares DynaFlow against several baselines, evaluating both dynamic feasibility and distributional similarity on two distinct datasets: a rich, strictly feasible dataset of quadruped locomotion and a challenging single-trajectory dataset from retargeted motion capture.
Our results demonstrate that DynaFlow consistently generates strictly feasible trajectories, even when trained on a physically inconsistent dataset, while remaining competitive in distributional similarity. To showcase its real-world applicability, we deploy action trajectories generated by DynaFlow on a physical Go1 quadruped robot. 
The robot successfully reproduces diverse gaits observed in the training data and executes long-horizon motions with high accuracy in challenging open-loop experiments, validating the precision and coherence of the generated actions.
Furthermore, we demonstrate its ability to translate infeasible retargeted motions into 
executable 
behaviors on hardware, bridging the gap between kinematic data and real-world execution.

The main contributions of this paper are as follows:
\begin{itemize}
\item 
\textbf{Dynamics-Embedded Generative Model}: 
We propose DynaFlow, a generative model that embeds a differentiable simulator into a flow matching framework to enforce dynamically consistent trajectory generation.

\item 
\textbf{Action-Free Learning via Analytical Gradients}:
Our method overcomes the scarcity of action data by leveraging analytical gradients to learn actions directly from state-only demonstrations.

\item 
\textbf{Real-World Validation}: 
We demonstrate the practical viability of our approach through hardware deployment on a physical quadruped robot, demonstrating successful real-world execution of the generated motions.
\end{itemize} 

\section{Related Works}

\begin{figure*}[t]
    \centering
    \includegraphics[width=1\textwidth]{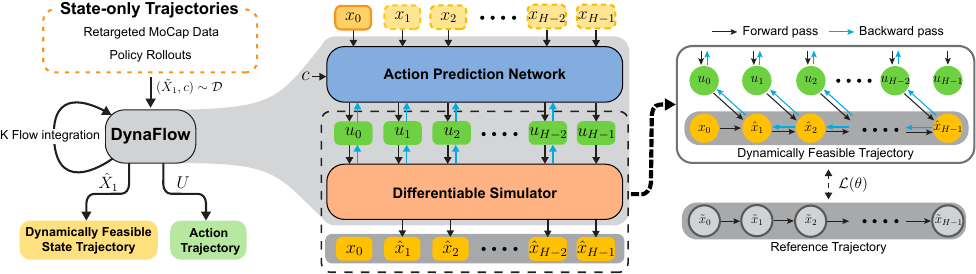}
    \captionsetup{font=small, skip=3pt}
    \caption{Schematic illustration of proposed framework. Given state-only demonstrations that are not guaranteed to be physically feasible, DynaFlow generates dynamically consistent trajectories while also inferring deployable action sequences. 
    The entire model is trained end-to-end using analytical gradients provided by the differentiable simulator.}
    \label{fig:DynaFlow}
\end{figure*}

\subsection{Generative Models for Control Tasks}

Generative models have been largely explored for physics-based character animation and robotic control. Prior works can be broadly categorized into two main paradigms.

The first paradigm is the hierarchical approach, where the generative model serves as a high-level state planner, while a separate low-level control module produces executable actions. This factorized design decouples planning and control, 
allowing the generative model to focus on state trajectory generation while other control modules handle deployment.
A key design choice in this paradigm is how to implement the control module.
For example, Decision Diffuser~\cite{decidiff_2022} uses ground-truth action labels to train an inverse dynamics network via supervised learning.
In the absence of action labels, one common approach is to pair the planner with a reinforcement learning-based tracking controller~\cite{closd_2024, robotmdm_2024}. Alternatively, model-based approaches combine generative models with optimization-based controllers, where diffusion models either generate intermediate subgoals that are tracked by Model Predictive Control (MPC)\cite{subgoal_2024} or provide initial guesses that are refined by trajectory optimization\cite{csvto_2023}. 
While effective, these approaches rely on auxiliary low-level tracking modules that are sensitive to out-of-distribution plans \cite{diffusecloc_2025} or require extensive task-specific tuning, limiting scalability.



The second paradigm is the direct generation approach, where the generative model itself infers executable control inputs. Models in this category either model action trajectories directly or jointly model state–action pairs. This line of work has shown notable success across diverse domains, including manipulation~\cite{diffusionpolicy_2023, adaflow_2024, pi0_2024}, legged locomotion\cite{diffuseloco_2024, birodiff_2024}, character control\cite{diffusecloc_2025, uniphys_2025, pdp_2024}, and humanoid whole-body control\cite{beyondmimic_2025}. While some works focus on generating action trajectories\cite{diffusionpolicy_2023, adaflow_2024, pi0_2024, diffuseloco_2024, pdp_2024}, others adopt joint formulations over state–action pairs~\cite{diffusecloc_2025} or couple states with latent action representations~\cite{uniphys_2025, beyondmimic_2025}. Such formulations alleviate the challenge of action sequence conditioning and further demonstrate how task-level guidance can steer generated actions for downstream tasks. Despite these advances, the reliance on ground-truth action datasets poses a practical limitation, as such data is costly to obtain and often requires training or fine-tuning of expert controllers for each task\cite{pdp_2024}.

Our framework departs from both paradigms by eliminating the need for ground-truth action labels or auxiliary tracking controllers, achieving scalable and physically consistent control directly from state-only demonstrations.

\subsection{Physics-Constrained Motion Generation}
There have been numerous attempts to incorporate physically plausible and dynamically consistent properties into generative models. One common approach to address this is to enforce feasibility through projection. PhysDiff~\cite{physdiff_2023} integrates a tracking controller into the denoising process, using it as a module to project each intermediate motion estimate onto a physically-plausible state. DDAT\cite{DDAT_2025} proposes several projection methods applicable to both state-only and state-action generation. In the state-only setting, it approximates the set of admissible states as a feasible region to project infeasible predictions. When ground-truth actions are available, it directly applies the action to predict an admissible next state, thereby ensuring dynamic consistency. DPCC\cite{dpcc_2024} integrates model-based projections into the denoising process to ensure explicit state and action constraints.

Another approach is to steer the generation process through guidance, which leverages gradients from a secondary function to refine outputs at inference time. The source of this gradient varies across methods. It can originate from explicit costs, such as a kinematic feasibility cost\cite{edmp_2024}, or from internal consistency measures\cite{refining_2023} that detects out-of-distribution states. To handle dynamic constraints with guidance where analytical gradients are often unavailable, RobotMDM~\cite{robotmdm_2024} addresses this by incorporating the critic of a tracking controller for an auxiliary loss, providing differentiable guidance toward physically consistent motions.

Our framework guarantees physical consistency by construction, embedding the simulator into the generative process, removing the need for external projection or guidance.

\subsection{Differentiable Physics}

The development of differentiable simulators\cite{brax_2021, mjx} has provided powerful tools for incorporating underlying physics directly into the learning process. By leveraging analytical gradients, these simulators enable sample-efficient training and provide interpretable learning signals from states.
Building on this capability, several works have proposed policy learning methods that utilize gradient 
from differentiable simulators to guide policy optimization\cite{pods_2021, shac_2022}. Another line of research leverages the ability to backpropagate from states to actions, reformulating motion or skill imitation as a state-matching problem. DiffMimic\cite{diffmimic_2023} demonstrated efficient learning of dynamic character control from a single demonstration, while ILD\cite{ild_2023} introduced a bidirectional state-matching objective that stabilizes training. In addition, differentiable simulation has been combined with state-diffusion planning to train latent action representations\cite{insactor_2023}, which are then employed as tracking controllers for language-guided character control.

Our framework directly embeds a differentiable simulator into the flow matching framework, optimizing actions
to match a target probability flow in the state space. This tight integration enables end-to-end learning of physically-grounded generative policy directly from state-only demonstrations.

\section{Background}\label{sec:background}

Flow Matching (FM)\cite{flowmatching} is a generative modeling framework that learns a time-dependent vector field $u(X,t)$ that transports samples from a simple base distribution $p_0(X)$ to a target data distribution $p_1(X)$ along paths connecting a prior sample $X_0 \sim p_0(X)$ and a data sample $X_1 \sim p_1(X)$. 

A common choice is the optimal transport path, which is given by the straight-line  $X_t = (1 - t){X}_0 + t {X}_1$,
where the model $u_\theta$ is trained to regress the true velocity field ${X}_1-{X}_0$ with the loss function
\begin{equation}
    \mathcal{L}_{\text{FM}}(\theta; {X}_0, {X}_1, t) = \left\| u_{\theta}({X}_t, t) - ({X}_1 - {X}_0) \right\|_2^2
\end{equation}
Once trained, new samples are generated by drawing $X_0 \sim p_0(X)$ and integrating the ordinary differential equation (ODE) $\dot{X}(t) = u_{\theta}(X(t), t)$ from $t=0$ to $t=1$.

More generally, in the affine conditional flow formulation, the interpolation is defined as $X_t = \alpha_t X_1 + \sigma_t X_0$,
where $\alpha_t$ and $\sigma_t$ are smooth schedules satisfying $\alpha_0 = 0$, $\alpha_1 = 1$, $\sigma_0 = 1$, $\sigma_1 = 0$, and $\dot{\alpha}_t>0 , \dot{\sigma}_t < 0$ for $t \in [0, 1]$. Then the marginal velocity field can be formulated as,
\[
u(X,t) \;=\; \frac{\dot{\sigma}_t}{\sigma_t} X 
+ \Bigg(\frac{\dot{\alpha}_t - \dot{\sigma}_t \alpha_t}{\sigma_t}\Bigg) \, \mathbb{E}[ {X}_1 |  {X}_t = X].
\]

This identity reveals that learning the posterior mean $\mathbb{E}[ {X}_1 |  {X}_t]$, also known as the $x_1$-prediction, is sufficient to define the entire flow velocity field~\cite{lipman2024flow}. 
In this work, we leverage this formulation by training our model to predict the posterior mean, as it provides a direct handle to enforce physical constraints on the generated trajectory.

\section{Dynamics-embedded Flow Matching}

Our proposed framework, DynaFlow, guarantees the dynamic feasibility of the generated trajectory by embedding a simulator as a deterministic mapping module at the output, which projects preceding outputs into the space of physically valid state trajectories, ensuring feasibility by construction.


For the remainder of this section, we denote the state trajectory by 
$X = x_{i:i+H}$ with $x_i \in \mathcal{X}$, 
and the action trajectory by 
$U = u_{i:i+H}$ with $u_i \in \mathcal{U}$ where $\mathcal{X}$ and $\mathcal{U}$ denote the state and action spaces. 
We use $\Phi$ to denote the differentiable rollout operator, mapping an initial state and an action trajectory to a full state trajectory, by recursively 
applying the one-step dynamics $x_{i+1} = f(x_i, u_i)$. 

\subsection{\textls[-22]{Differentiable Physics as a Dynamics-Aware Embedding Layer}}

The core idea of DynaFlow is to structure the posterior mean to explicitly respect system dynamics. To this end, we factorize the prediction module into two components: 

\begin{itemize}
    \item \textbf{Action Prediction Network.}  A network $D_\theta$ predicts a sequence of future actions $U$ conditioned on the current noisy state trajectory $X_t$, 
    conditioning $c$ including observations, command signals, and the time step $t$:
    \begin{equation}
    \scalebox{0.95}{$
        \hat{U} = D_\theta(X_t, c, t)
    $}
    \end{equation}
    
    \item \textbf{Differentiable Dynamics.} A differentiable rollout operator $\Phi$ serves as a deterministic mapping from the initial state $x_0$ together with the predicted actions $\hat{U}$ into a physically realizable state trajectory:
    \begin{equation}
    \scalebox{0.95}{$
        \hat{X}_1 = \Phi(x_0, \hat{U})
        \label{eq:action_prediction}
    $}
    \end{equation}
    
\end{itemize}
This factorization ensures that the resulting prediction $\hat{X}_1$ is dynamically feasible, by construction.

\subsection{Training Objective}

We train the action prediction network $D_\theta$ with the Conditional Matching (CM) loss, a variant of the Flow Matching objective that trains the model to predict the posterior mean $\mathbb{E}[X_1 | X_t]$. 
The objective is to minimize the expected discrepancy between the model's prediction and the ground-truth trajectory drawn from the target distribution.

\vspace{-0.5em}
\begin{equation}
\scalebox{0.95}{$
    \mathcal{L}(\theta) = 
    \mathbb{E}_{t, X_0, X_1} \left[ 
        \left\| 
            W \odot ( \hat{X}_1 - X_1 ) 
        \right\|_2^2 
    \right],
    \label{eq:loss_fn}
$}
\end{equation}

Here, the expectation is over the time $t \sim \mathcal{U}(0,1)$, the base distribution $X_0 \sim p_0$, and the target data distribution represented by the random variable $X_1 \sim p_1$. $X_t$ denotes their linear interpolation, and $W$ is a weighting mask applied element-wise to balance contributions across state dimensions and timesteps.
In practice, we approximate this expectation
by sampling an expert trajectory $\tilde{X}_1 \sim \mathcal{D}$ and generate a corresponding prediction $\hat{X}_1 = \Phi(x_0, D_\theta(X_t, c, t))$. 

\subsection{Velocity Field Formulation}

Our training objective is theoretically grounded in the identity of affine conditional flows, connecting the velocity field to the posterior mean.
In the special case of the optimal transport path, where $\alpha_t = t$ and $\sigma_t = 1 - t$, we obtain:
\vspace{-0.2em}
\begin{equation}
    u_\theta(X_t, t) = \frac{1}{1 - t} \left( \mathbb{E}[X_1|X_t] - X_t \right)
    \label{eq:velocity_field}
\end{equation}

The posterior mean $\mathbb{E}[X_1 | X_t]$ is obtained by applying the rollout operator $\Phi$ to the predicted action sequence. Since $\Phi$ is deterministic, this conditional expectation can be written as a marginalization over the action distribution:
\vspace{-0.2em}
\begin{equation}
\scalebox{0.95}{$
\begin{aligned}
\mathbb{E}[X_1 | X_t] 
&= \int \Phi(x_0, U) \, p_\theta(U | X_t) \, dU \\
&= \mathbb{E}_{U \sim p_\theta(\cdot \mid X_t)} \left[ \Phi(x_0, U) \right]
\end{aligned}
\label{eq:posterior_mean}
$}
\end{equation}


As the posterior mean is computed through the rollout operator, the learned velocity field naturally inherits a dynamics-aware structure. The transport to the target distribution is then obtained by integrating the corresponding ODE.
In practice, we discretize the time domain $t \in [0, 1]$ using a first-order explicit Euler solver,
$
X_{t+\Delta t} = X_t + \Delta t \cdot u_\theta(X_t, t),
$
where $\Delta t$ is the integration step size. We find that even a single step with $\Delta t=1$ is sufficient to produce high-quality predictions for control tasks, 
which allows real-time inference on hardware. 

\begin{algorithm}
\caption{DynaFlow (Training)}
\label{alg:DynaFlow}
\begin{algorithmic}[1]
\Require Expert dataset $\mathcal{D}$, differentiable simulator $\Phi$, action prediction network $D_\theta$
\Ensure Trained parameters $\theta$
\While{not converged}
        \State Sample $(\tilde{X}_1, c) \sim \mathcal{D}$, $X_0 \sim p_0(X)$, $t \sim \mathcal{U}(0,1)$
    \State Interpolate trajectory: $X_t = (1 - t)X_0 + t \tilde{X}_1$
    \State Predict actions: $\hat{U} \gets D_\theta({X}_t, c, t)$
    \State Rollout trajectory: $\hat{X}_1 = \Phi(x_0, \hat{U})$
    \State Compute loss with \eqref{eq:loss_fn}
    \State Update $\theta$ via gradient descent
\EndWhile
\end{algorithmic}
\end{algorithm}

\begin{algorithm}
\caption{DynaFlow (Inference)}
\label{alg:inference_DynaFlow}
\begin{algorithmic}[1]
\Require trained action prediction network $D_\theta$, differentiable rollout module $\Phi$, initial state $x_0$, conditioning $c$, integration step size $\Delta t$
\Ensure dynamically feasible trajectory $X_1$
\vspace{0.5em}

\State Initialize $X_t \gets X_0 \sim p_0(X)$
\For{$t = 0$ \textbf{to} $1$}

\State Predict actions: $\hat{U} \gets D_\theta(X_t, c, t)$ 
\State Predict ${\mathbb{E}}[X_1 | X_t] \gets \Phi(x_0, \hat{U})$
\State Compute velocity field $u_\theta(X_t, t)$ with \eqref{eq:velocity_field}
\State Update $X_t \gets X_t + \Delta t \cdot u_\theta(X_t, t)$
\State $t \gets t + \Delta t$

\EndFor
\State \Return $X_1 \gets X_t$
\end{algorithmic}
\end{algorithm}

\begin{figure*}[t]
    \centering
    \includegraphics[width=0.99\linewidth]{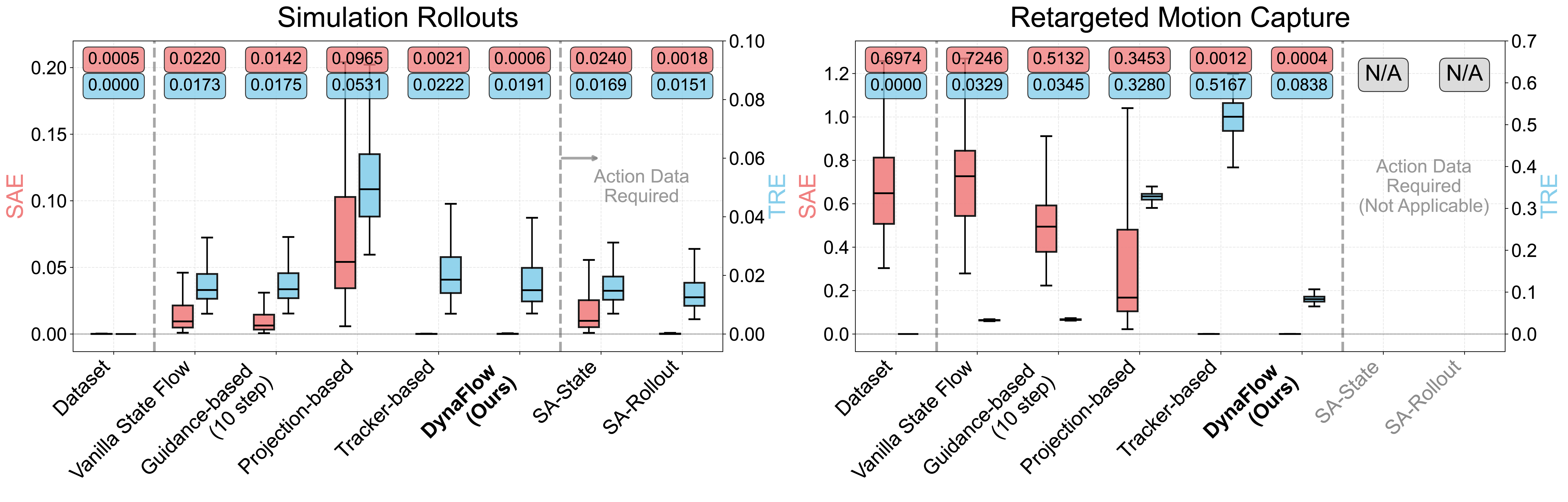}
    \captionsetup{font=small, skip=3pt}
    \caption{Comparative analysis of physical consistency and distributional similarity. The SAE (red) and TRE (blue) for each method on two distinct datasets (Sec.~\ref{ssec:dataset}) are summarized in the box plot. The mean value for each metric is annotated above the corresponding box plots. 
    Notably, DynaFlow produces physically consistent trajectories while preserving high fidelity to the original motion.
    }
    \label{fig:quantitative_analysis}
\end{figure*}

\subsection{Datasets}\label{ssec:dataset}
We use two distinct datasets to evaluate performance under both controlled and challenging conditions. The first, \emph{Simulation Rollouts Dataset}, comprises about 12,000 5-second trot and bound trajectories generated by rolling out a pre-trained policy~\cite{wtw} under various velocity commands. 
To establish a strictly feasible baseline for quantitative analysis, we 
re-generated the \emph{Simulation Rollouts Dataset} in MuJoCo~\cite{mujoco} under identical conditions, yielding dynamically feasible state–action pairs for controlled evaluation.
The second, \emph{Retargeted Motion Capture Dataset}, is a more challenging state-only dataset consisting of a single 2.54-second galloping trajectory. The motion was created by adapting German Shepherd motion capture data~\cite{gallopmocap2018} to a Go1 robot skeleton, following the retargeting procedure of Peng et al.~\cite{learning2020peng}. As kinematic retargeting does not incorporate dynamic properties or enforce actuation constraints, the resulting trajectory is often dynamically infeasible, exhibiting physical artifacts such as ground penetration. 
This makes it a demanding benchmark of model robustness and reflects real-world scenarios where only sparse kinematic data are available.
The contrasting properties of these datasets are summarized in Table~\ref{tab:datasets}.

\begin{table}[t]
\centering
\footnotesize
\caption{Evaluation dataset characteristics.}
\renewcommand{\arraystretch}{1.2}
\begin{tabular}{@{}>{\centering\arraybackslash}m{2.5cm}ccc@{}}
\hline
Dataset & Data & Size & Feasibility \rule{0pt}{2.5ex}\\
\hline
\emph{Simulation Rollouts} & State+Action & 12k traj. (16.7h) & O \\
\emph{Retargeted MoCap} & State only & 1 traj. (2.54s) & X \\
\hline
\end{tabular}
\label{tab:datasets}
\end{table}



\subsection{Implementation Details}

Our framework is implemented in JAX\cite{jax}, with MuJoCo XLA\cite{mjx} serving as the differentiable simulation backend.

Our model is trained on trajectories with a fixed horizon of $H{=}16$. The state vector $x_i \in \mathbb{R}^{37}$ encompasses the system's generalized coordinates and velocities. The action vector $u_i \in \mathbb{R}^{12}$ is represented as a residual joint position target relative to a nominal joint position, normalized per joint.

We use a 1D Diffusion Transformer (DiT) backbone with three transformer blocks (10.3M parameters),
conditioned with sinusoidal positional encodings over the horizon, MLP-based timestep embeddings, and task-specific attributes $c = [\, g, v, \omega, q, \dot{q}, z_{\mathrm{gait}}, v_x^{\mathrm{cmd}}, v_y^{\mathrm{cmd}},    \omega_z^{\mathrm{cmd}}]$. This vector includes gravity, base velocities, joint states, gait mode, and velocity commands. 
We train with Adam\cite{adam} optimizer with a learning rate of $2{\times}10^{-4}$, and apply Exponential Moving Averaging (EMA) with decay $0.995$ for 25k steps ($\approx$ 3h) on a single NVIDIA RTX 4090 GPU with a batch size of 1024.



\section{Results}

\subsection{Comparative Analysis on Physical Consistency and Distributional Similarity}
To quantitatively evaluate the effectiveness of DynaFlow, we conducted a comparative analysis against several baselines. 

\begin{figure}[t]
    \centering
    \includegraphics[width=0.98\linewidth]{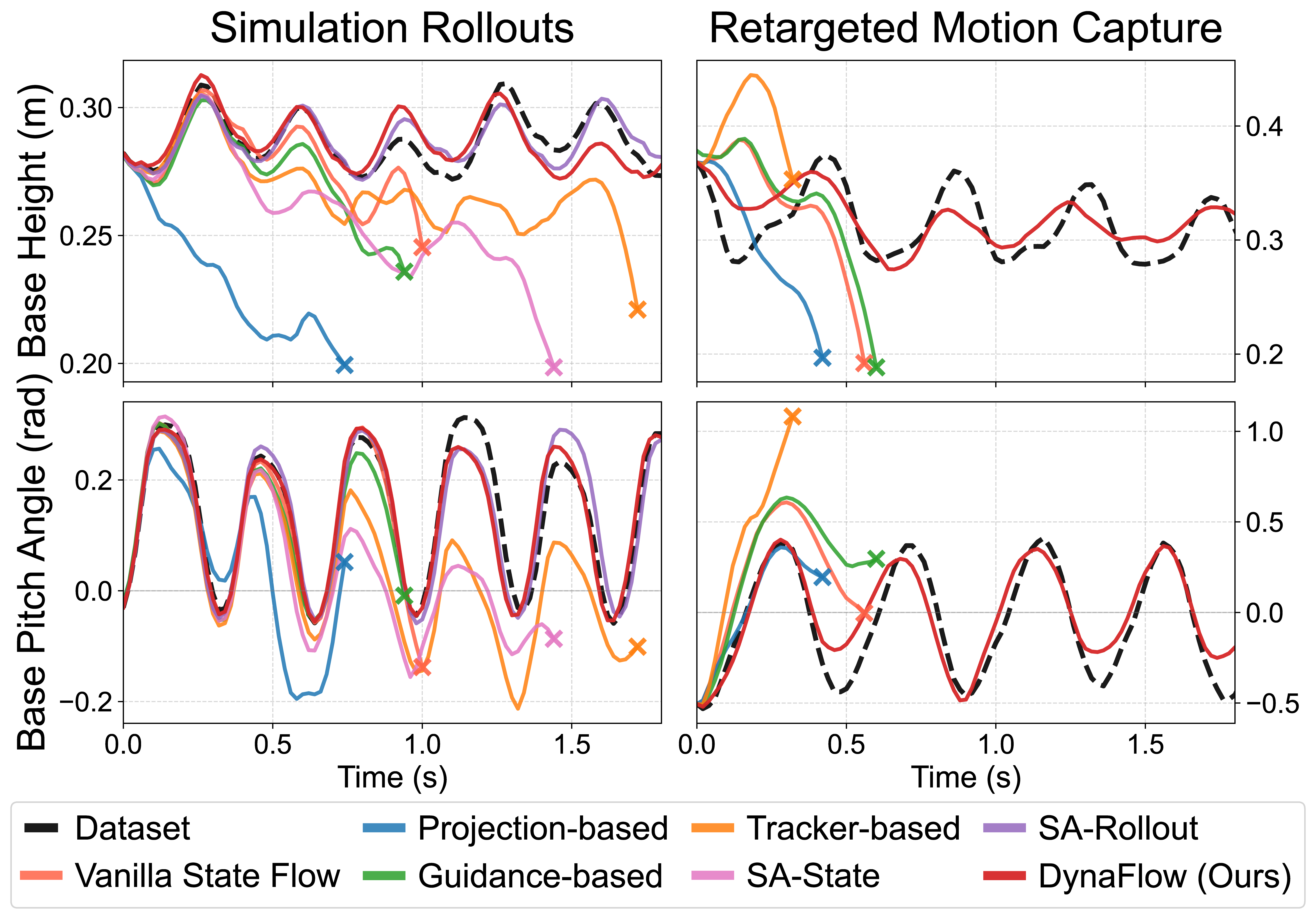}
    \captionsetup{font=small, skip=3pt}
    \caption{Comparison of tracking performance on the (a) Simulation Rollouts and (b) Retargeted Motion Capture datasets. Each plot shows the resulting state trajectory (base height and pitch angle) after using a numerical inverse dynamics solver to track the original plan from each method. Trajectories are terminated and marked with an `x' upon failure (body height $<$ 0.2~m or the vertical component of the base's z-axis $<$ 0.5). Most baselines produce untrackable plans that fail due to either the accumulation of physical errors (SAE) or inherent plan instability (high TRE). In contrast, only DynaFlow consistently generates trackable and stable trajectories.}
    \label{fig:qualitative_analysis}
\end{figure}

\subsubsection{Evaluation Metrics}
Our evaluation focuses on two primary criteria: \textit{Physical Consistency}, which measures adherence to the system's dynamic constraints, and \textit{Distributional Similarity}, which assesses how well the generated motions maintain fidelity to the original data distribution.

To measure physical consistency, we use the Statewise Admissibility Error (SAE), introduced in DDAT~\cite{DDAT_2025}. For a given transition from $x_i$ to ${x}_{i+1}$, the SAE is the Euclidean distance between ${x}_{i+1}$ and the closest reachable state under the system dynamics $f(x_i, u) $ for any valid action $u \in \mathcal{U}$. The optimal action that achieves this closest state is obtained by solving the following inverse dynamics problem:

\vspace{-1.0em}
\begin{align}
\scalebox{0.95}{$
\begin{aligned}
\text{ID}(x_i, x_{i+1}) &:= \arg\min_{u \in \mathcal{U}} \| x_{i+1} - f(x_i, u) \|, \\[2pt]
\text{SAE}(x_i, x_{i+1}) &:= \| x_{i+1} - f\!\left(x_i, \text{ID}(x_i, \tilde{x}_{i+1})\right)\|.
\end{aligned}
\label{eq:inverse_dynamics_SAE}
$}
\end{align}



We solve this inverse dynamics problem numerically using gradient descent, taking advantage of the differentiability of $f$.
A lower SAE indicates higher physical consistency, with a score of zero implying a dynamically feasible transition.



To evaluate distributional similarity, we employ the Trajectory Reconstruction Error (TRE), which measures the mean squared error between a reference trajectory $\tilde{X}$ and a generated trajectory $\hat{X}$, with
the same initial state $x_0$.

To establish a meaningful correspondence, we condition the generation on the initial state $x_0$ and fix the initial Gaussian sample across baselines. Consequently, a lower TRE signifies higher fidelity to the data distribution, acting as an upper bound on the 2-Wasserstein distance.



\subsubsection{Baselines}
We compare DynaFlow against the following state-only baselines:
\begin{itemize}
    \item \textbf{Vanilla State Flow}: 
    A standard flow matching model trained directly on state trajectories. 
    \item \textbf{Guidance-based}: 
    A self-implemented baseline that employs classifier guidance to promote physical consistency. It uses a separately trained inverse dynamics model to form a differentiable SAE function, whose gradients guide the generative process towards dynamically feasible trajectories during inference.
    \item \textbf{Projection-based}: 
    A post-processing method where generated state trajectories are projected onto a polytope that approximates the reachable set, following the state-only projection method from DDAT~\cite{DDAT_2025}. For computational efficiency, instead of sampling all 4,096 extremum points, we approximate the reachable set by using 512 extremum points and 512 randomly sampled points as our baseline. 
     
     \item \textbf{Tracker-based}: A two-stage method that first generates a state trajectory with a vanilla model and then uses a tracker (with the same inverse dynamics model as Guidance-based) to sequentially infer and execute actions, producing the final rollout~\cite{physdiff_2023}.
     
\end{itemize}

\noindent For completeness, we also compare against a model trained with ground-truth action data.
\begin{itemize}
    \item \textbf{State Action Flow (SA)}: A flow matching model trained on paired state-action data. We evaluate two variants: \textit{SA-State}, its directly predicted state trajectories, and \textit{SA-Rollout}, the state trajectories produced by rolling out its predicted action trajectories.
\end{itemize}

\subsubsection{Results}

The quantitative results of our comparative analysis are illustrated in Fig.~\ref{fig:quantitative_analysis}. The overall results demonstrate that DynaFlow uniquely achieves strict physical consistency while preserving the data distribution. 
To understand the practical implications of these metrics, we also visualized the control performance by tracking each generated trajectory with a numerical inverse dynamics solver, as shown in Fig.~\ref{fig:qualitative_analysis}.

On the dynamically feasible \emph{Simulation Rollouts} dataset, the box plot shows that DynaFlow achieves near-zero SAE, similar to SA-Rollout, which benefits from ground-truth action labels. We note that this minimal non-zero SAE reflects the precision limit of the numerical inverse dynamics solver for evaluation rather than a model failure. In contrast, other state-predictive baselines like Vanilla State Flow and SA-State introduce noticeable feasibility errors, revealing that generative models can violate physical laws even when trained on ideal data. In terms of distributional similarity, DynaFlow attains a TRE comparable to the strongest baselines. The tracking experiment in Fig.~\ref{fig:qualitative_analysis} clearly demonstrates the consequence of these metrics: even small physical errors from methods like Vanilla State Flow accumulate, causing the controller to fail. Conversely, the strict physical consistency
of DynaFlow's trajectories enables 
stable 
locomotion. 

The performance gap becomes more pronounced on the physically inconsistent \emph{Retargeted Motion Capture} dataset. This scenario invalidates action-supervised methods like State Action Flow. As seen in the box plot, 
the dataset exhibits
a high intrinsic SAE, which methods like Vanilla State Flow simply replicate. While corrective approaches (Guidance-based, Projection-based) offer partial improvements, they fail to fully resolve the infeasibility. The Tracker-based method successfully reduces the SAE to near-zero, but at the cost of a drastically increased TRE, indicating a significant deviation from the reference motion's structure. In this context, DynaFlow is the only method that maintains both low SAE and low TRE simultaneously. The tracking results starkly reflect this. The high SAE of most baselines makes their plans untrackable from the start. 
Notably, even the Tracker-based method fails, as its physically plausible but structurally incoherent plan (high TRE) is too unstable to follow. Only DynaFlow, by satisfying both physical consistency and distributional similarity, produces a plan that can be successfully tracked, preserving 
motion style.

These results underscore a fundamental requirement for generative models in robotics: a significant trade-off exists between achieving physical consistency and maintaining distributional similarity. A generated plan is only useful if it is both physically possible (low SAE) and coherent (low TRE). DynaFlow, by integrating the differentiable simulator directly into its structure, effectively resolves this trade-off. Its design achieves strict physical consistency by construction, while incurring only a minimal loss in distributional similarity.

\subsection{Real-World Experiments}
\begin{figure}[t]
    \centering    \includegraphics[width=1.0\linewidth]{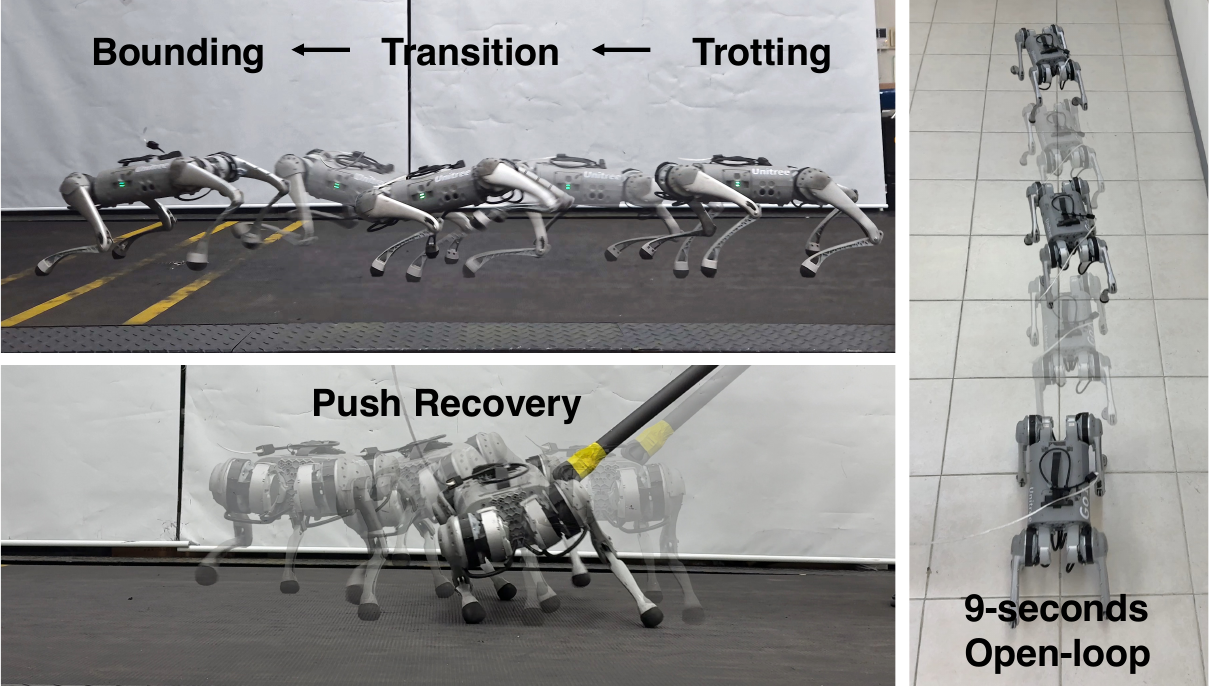}
    \captionsetup{font=small, skip=3pt}
    \caption{Performance validation of DynaFlow on the real-world experiment. The robot demonstrates a smooth transition between gaits, robust recovery from an external push, and successful long-horizon open-loop locomotion without any replanning.}
    \label{fig:basic locomotion}
\end{figure}

To validate the motions learned by DynaFlow, we conducted
real-world experiments on a Unitree Go1 quadruped robot.
For all experiments, we set the integration step size to $\Delta t = 1$ to 
minimize latency for real-time hardware execution.

As shown in Fig.~\ref{fig:basic locomotion}, we 
evaluated DynaFlow on trotting and bounding motions learned from the \emph{Simulation Rollouts} dataset. For deployment, we adopted a receding-horizon 
scheme with a 10~Hz replanning rate, executing the first five actions of each 
plan. Under this setup, the robot successfully executed locomotion at speeds up to 2.0~m/s for each gait, consistent with the velocity range covered in the training data. 
Although explicit gait transitions were absent from the training data,
the robot exhibited smooth and stable switching between gaits at a forward speed of 1.5~m/s. 
To further assess DynaFlow’s long-horizon feasibility and coherence,
we conducted a challenging open-loop experiment by generating a 450-step (9-second) action trajectory through iterative sampling. 
We repeatedly used the final state of a predicted trajectory segment as the initial state for generating the next segment, sequentially chaining them together. Using this trajectory, the robot successfully traversed a 4~m forward path within a 1~m-wide corridor using both gaits, 
without any replanning. This highlights DynaFlow’s ability to maintain long-term consistency, enabling entire sequences to run on real hardware in open-loop.

We further evaluated DynaFlow on a dynamically infeasible kinematic demonstration using the \emph{Retargeted Motion Capture} dataset. In this experiment, 
a Vicon motion capture system provided base-frame linear velocity for observations.

\begin{figure}[t]
    \centering
    \includegraphics[width=1.0\linewidth]{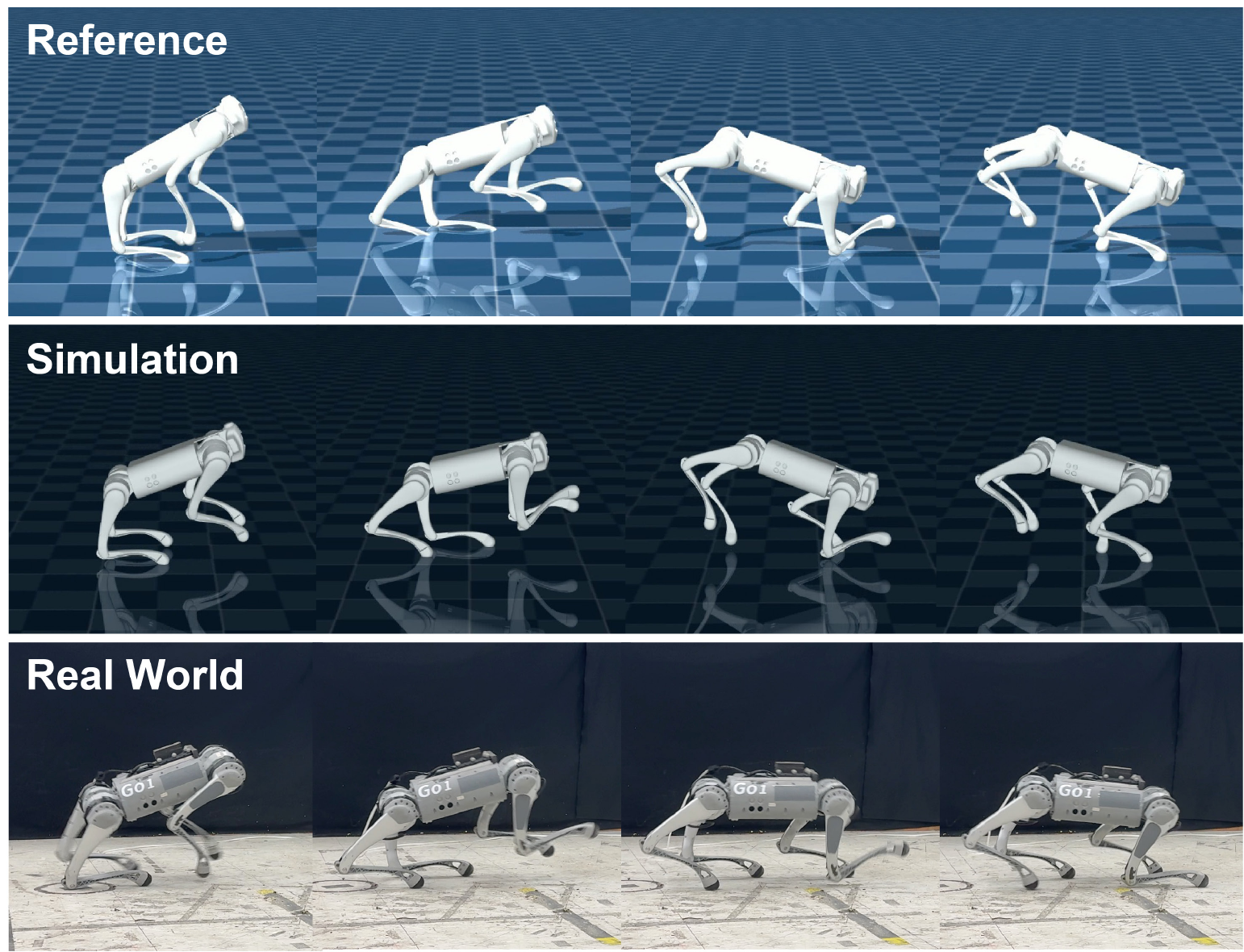}
    \captionsetup{font=small, skip=3pt}
    \caption{Transfer of an infeasible motion to real-world execution. DynaFlow translates a dynamically infeasible, kinematically retargeted German Shepherd gallop into a feasible trajectory and successfully executes it in simulation and on the Go1 robot.}
    \label{fig:gallop transfer}
    \vspace{-1em}
\end{figure}

The results highlight DynaFlow’s ability to transform infeasible reference trajectories into physically grounded behaviors while preserving 
stylistic characteristics. As shown in Fig.~\ref{fig:gallop transfer}, the model 
transforms the inconsistent kinematic demonstration into a dynamically feasible trajectory in simulation,
which is then executed on the real robot without 
fine-tuning or adaptation, resulting in a stable galloping motion.

\begin{figure}[t]
    \centering
    \includegraphics[width=1.00\linewidth]{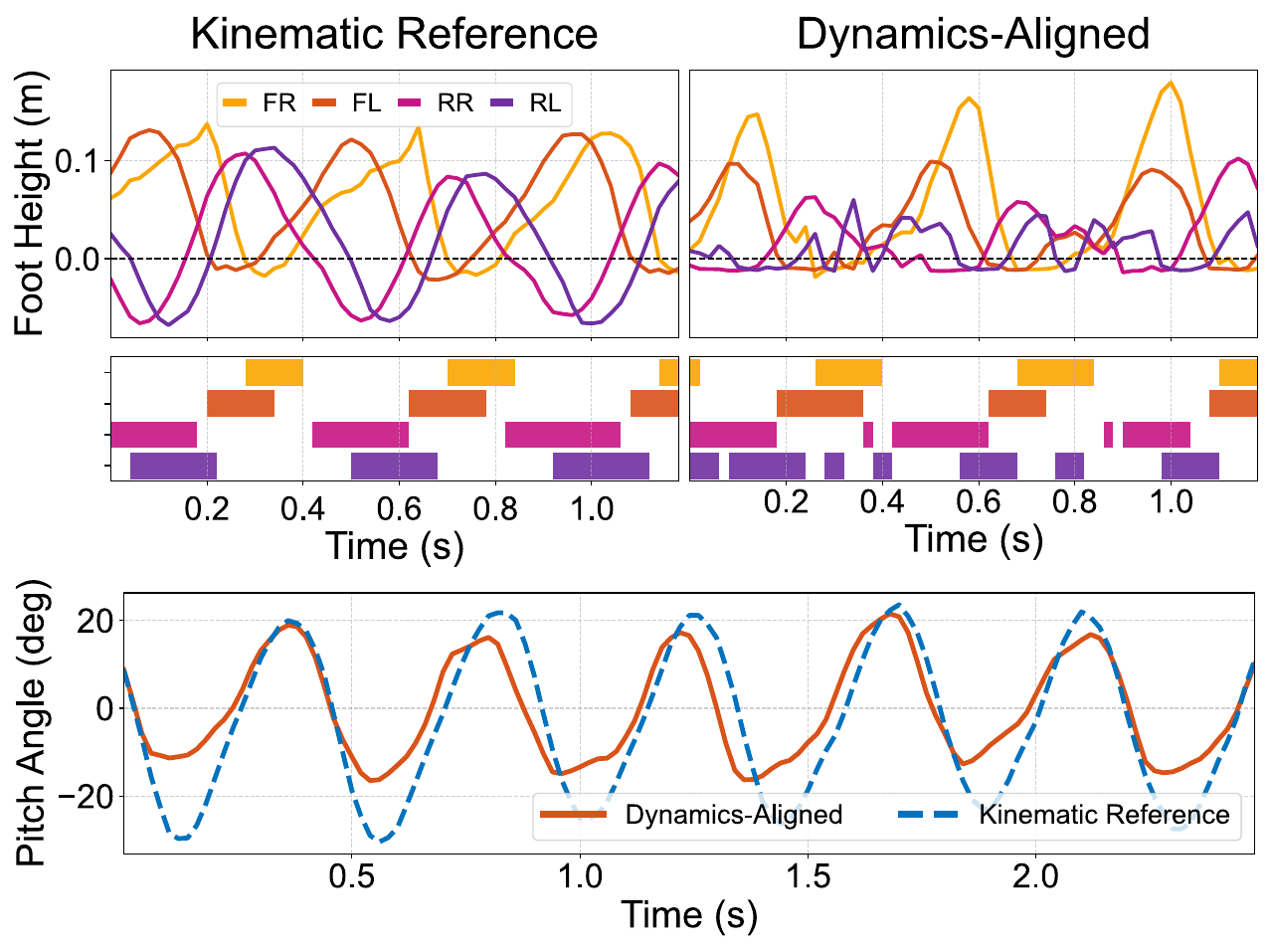}
    \captionsetup{font=small, skip=3pt}
    \caption{Simulation analysis of reference motion and DynaFlow rollout, showing reduced foot–ground penetrations and base pitch oscillations while preserving the characteristic galloping style.}
    \label{fig:infeasibility}
\end{figure}

Figure \ref{fig:infeasibility} illustrates how DynaFlow resolves the physical inconsistencies in the reference motion, which would otherwise prevent hardware deployment. It removes foot-ground penetrations (up to 5~cm) and reduces the base-pitch oscillation from nearly $30^{\circ}$ to about $20^{\circ}$; the small residual penetration visible in the plot reflects the simulator’s soft-contact model and solver tolerances.
Crucially, these corrections preserve the motion’s core stylistic features. The characteristic galloping contact sequence is maintained, indicating that physical consistency is achieved without compromising style. The full demonstrations of all hardware experiments are provided in the supplementary video.

\subsection{Robustness to External Disturbances}

To evaluate the robustness of our model, we tested its ability to withstand external disturbances not encountered during training. In 1,000 simulated trials, the robot was commanded to trot forward at a target velocity of 1.0 m/s for 4 s, while a random horizontal perturbation force of up to 50 N was applied during $t\in[1, 2]s$ in each trial. We compared DynaFlow against the SA-Rollout baseline, deploying both with a 50~Hz receding horizon control scheme. The evaluation was conducted across a range of replanning intervals (window sizes) and flow integration steps. A trial was considered a failure if the body height dropped below 0.15~m or the vertical component of the z-axis of its rotation matrix fell below 0.5.

\begin{figure}[t]
    \centering
    \includegraphics[width=1.00\linewidth]{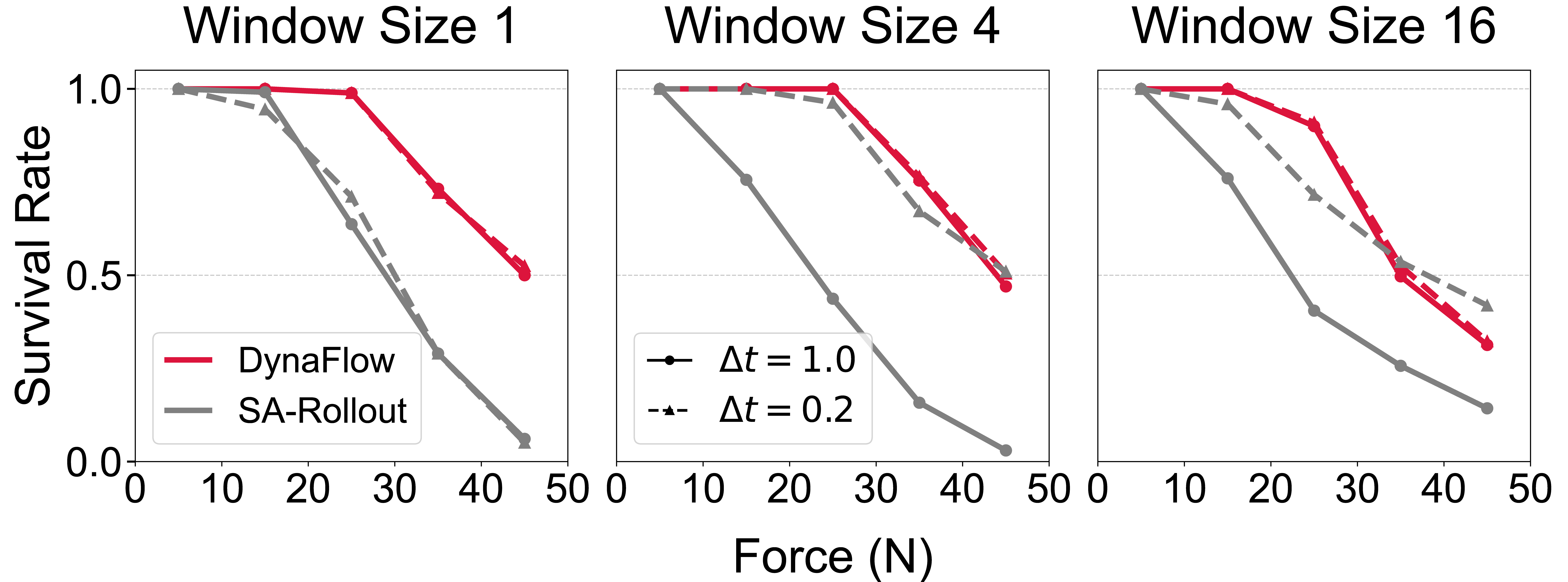}
    \captionsetup{font=small, skip=3pt}
    \caption{Survival rate under increasing external disturbances in simulation. DynaFlow consistently outperforms the SA-Rollout baseline and maintains high performance across configurations.}
    \vspace{-1em}
    \label{fig:recovery}
\end{figure}

As illustrated in Fig.~\ref{fig:recovery}, DynaFlow 
achieves a higher or comparable survival rate than the SA-Rollout baseline across most configurations. Furthermore, DynaFlow's performance remains remarkably consistent across different window sizes and integration timesteps. We attribute this enhanced robustness to the rich learning signal provided by the differentiable simulator. Unlike SA-Rollout, 
which directly imitates action sequences,
DynaFlow receives gradients through the dynamics model during its training. This process implicitly encourages the model to learn not just the actions to follow a trajectory, but also the corrective actions needed to remain on a stable path, thus improving its resilience to unforeseen perturbations.

\section{Discussion}

In this paper, we introduced DynaFlow, a novel generative framework that addresses the challenges of physical inconsistency and action data reliance in robot control. By embedding a differentiable simulator into a flow matching model, DynaFlow guarantees dynamically feasible motions
by construction. Its differentiability enables end-to-end training on state-only demonstrations, 
inferring actions to reconstruct a given motion. 
Quantitative results show
that DynaFlow produces strictly feasible motions even from physically inconsistent data while preserving distributional fidelity. 
Our real-world deployments further showcased its practical effectiveness,
successfully executing diverse 
locomotions under both closed-loop and open-loop conditions.

\textcolor{black}{
Future directions include several extensions.
First, while our current implementation utilizes a fixed horizon of $H=16$, analyzing 
gradient stability
over extended horizons will be crucial for 
long-term tasks.
Second, expanding to more complex platforms with rich dynamically infeasible datasets, such as humanoids, will offer a more rigorous evaluation of the framework’s robustness.
Finally, while this work primarily focuses on locomotion, extending DynaFlow to dynamic loco-manipulation and contact-rich object interaction 
is a promising direction.
Our supplementary video includes a preliminary demonstration of a ball manipulation task.
}








\bibliographystyle{IEEEtran}

\bibliography{Format/refs}

\end{document}